\documentclass{Interspeech}

\interspeechcameraready

\title{Evaluating ASR robustness to spontaneous speech errors: A study of WhisperX using a Speech Error Database}

\author[affiliation={1}]{John}{Alderete}
\author[affiliation={2}]{Macarious Kin Fung}{Hui}
\author[affiliation={2}]{Aanchan}{Mohan}

\affiliation{Linguistics and Cognitive Science}{Simon Fraser University}{Canada}
\affiliation{Khoury College of Computer Sciences}{Northeastern University, Vancouver, BC}{Canada}
\email{alderete@sfu.ca, hui.mac@northeastern.edu, aa.mohan@northeastern.edu}
\keywords{automatic speech recognition, model evaluation, benchmarks, model training, speech errors, psycholinguistics, contextual sensitivity}

\usepackage{comment}
\usepackage{booktabs}  
\usepackage{array}     
\usepackage{ragged2e}  
\usepackage{graphicx}   
\usepackage{caption}    
\usepackage{subcaption} 
\usepackage{float}      
\usepackage{placeins}   
\usepackage{tipa}
\begin{document}

\maketitle

\begin{abstract}
    
The Simon Fraser University Speech Error Database (SFUSED) is a public data collection developed for linguistic and psycholinguistic research. Here we demonstrate how its design and annotations can be used to test and evaluate speech recognition models. The database comprises systematically annotated speech errors from spontaneous English speech, with each error tagged for intended and actual error productions. The annotation schema incorporates multiple classificatory dimensions that are of some value to model assessment, including linguistic hierarchical level, contextual sensitivity, degraded words, word corrections, and both word-level and syllable-level error positioning. To assess the value of these classificatory variables, we evaluated the transcription accuracy of WhisperX across 5,300 documented word and phonological errors. This analysis demonstrates the database's effectiveness as a diagnostic tool for ASR system performance. 
\end{abstract}

\section{Introduction}

Automatic speech recognition (ASR) technology relies heavily on high-quality datasets for both training and evaluation. Current datasets effectively capture many dimensions of speech complexity, including variations in speech style (read versus spontaneous), accent, speaker characteristics, noise conditions, and linguistic genres \cite{alharbi2021automatic, graham2024evaluating}. However, these datasets operate under an implicit assumption that transcriptions are free of human-produced errors—an assumption that overlooks a fundamental characteristic of natural speech. Research has consistently shown that spontaneous speech contains speech errors, or slips of the tongue, which represent deviations from intended speech plans \cite{fromkin1971non, shattuck1979speech}. Consequently, ASR systems are sometimes trained and evaluated using these human-produced errors as ground truth, potentially impacting model development and assessment.

Speech errors occur approximately once or twice a minute in natural conversation \cite{alderete2019investigating}, making them a significant factor in spoken language processing. At typical speaking rates of 150 words per minute \cite{maclay1959hesitation}, speech errors affect approximately 1\% of utterances. Recent research has shown that training data containing even small amounts of human transcription errors leads to a two-fold increase in Word Error Rate (WER) \cite{gao2023human, sun2023htec}, indicating the importance of accurate speech error detection in ASR training data. Given the current focus on WER reduction in end-to-end models \cite{prabhavalkar2023end}, addressing the problem of speech errors may represent low-hanging fruit for model improvement.

Beyond model performance, speech errors provide a valuable test case for ASR robustness. While ASR research has traditionally focused on accommodating linguistic variation and degraded speech signals \cite{li2014overview}, speech errors present unique challenges. These include mid-word interruptions, phoneme additions, deletions, and substitutions, as well as word substitutions that lack explicit phonetic cues to the speaker's intended utterance. Evaluating models against these natural distortions provides insight into their capacity to handle extreme cases of signal variation.

Speech errors also exhibit systematic patterns that provide unique testing opportunities for ASR models. For example, most sound errors are contextual in the sense that they involve substitutions of sounds that occur in the neighboring linguistic context \cite{stemberger1982lexicon}. This context dependency provides an opportunity to investigate the local context encoding and forward prediction capabilities in ASR models, particularly in streaming applications where future context is unavailable. Understanding ASR model performance on speech errors also has broader implications for processing impaired speech, such as aphasic speech, where error rates are substantially higher \cite{dell1997lexical}. Advances in error transcription could therefore directly benefit assistive speech technologies.

\begin{figure}[t]
  \centering
  \includegraphics[width=\linewidth]{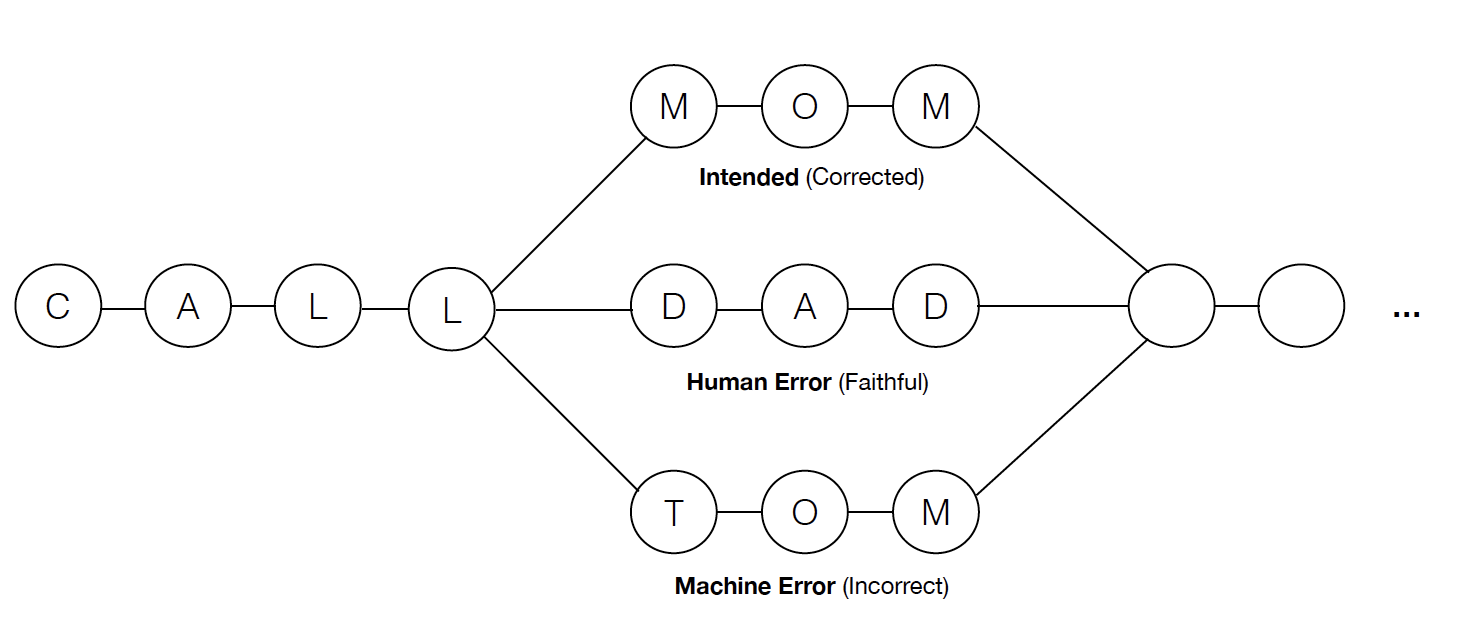}
  \caption{Recognition lattice with speech errors.}
  \label{fig:rec_lattice}
\end{figure}

Finally, labeled speech error datasets enable the construction of recognition lattices that incorporate both the produced error and the intended target word. Figure~\ref{fig:rec_lattice} illustrates this lattice structure with the error word \textit{Dad} and the intended word \textit{Mom}. The resulting lattice supports three potential transcription paths: faithful error reproduction (\textit{Dad}), error correction to the intended word (\textit{Mom}), or alternative high-probability candidates (\textit{Tom}). While traditional sequence discrimination models utilizing the lower two lattice paths have achieved significant WER reductions \cite{kingsbury2009lattice, povey2008boosted}, the inclusion of the error path creates new optimization possibilities. Models can be tuned to prioritize either error correction by weighting intended word paths more heavily, or verbatim transcription by emphasizing actually produced error words.

To demonstrate this approach, we analyzed WhisperX \cite{bain2023whisperx} transcriptions of 5,300 speech errors from SFUSED English \cite{alderete2019simon}, the largest corpus of spontaneous speech errors derived from audio recordings. Our analysis examines not only overall transcription accuracy but also explores how error contexts and conditions affect model performance, leveraging SFUSED English's rich cross-classification of linguistic variables.

\section{Methods}

\subsection{SFUSED English}
The Simon Fraser University Speech Error Database (SFUSED) English contains 10,000 labeled speech errors recorded across 360 hours of spontaneous speech from third-party podcast series. The source material includes podcasts like \textit{The Astronomy Podcast} and \textit{Rooster Teeth} that were selected on the basis of the quantity of unscripted speech, gender balance, and high production quality. Each podcast episode is between 30-60 minutes long. The SFUSED English documentation describes speaker characteristics, data quality, error mark-up, workflows, and access to the audio \cite{alderete2019simon}.\footnote{https://osf.io/8c9rg/} The audio data comes with marked up speech errors, but it does not come with human ground-truth transcriptions. As a result, we are unable to provide a baseline Word Error Rate.

We extracted 5,300 speech errors from SFUSED English, cross-classified by several experimental conditions relevant to ASR testing.\footnote{The complete dataset and classification script described below are available at: https://osf.io/6x4se/.} As shown in Table \ref{tab:errors}, they differ in error type: word substitution errors (2,4) involve a complete replacement of the intended word for a different word, whereas sound errors (1,3) involve slight mispronunciations of an intended word, including sound substitutions, additions, and deletions. Word errors present a greater challenge for ASR models than sound errors because the intended and error words are entirely different and often there is no phonetic evidence for the intended word. 

In addition to labeled error words, SFUSED English's analysis also includes a variable for the intended word, which is inferred either from corrected errors or the logic of the conversation (as in \textit{password} in (4)). The SFUSED English documentation states that intended words can be ascertained in 98\% of cases, and the remaining examples are labeled for low-confidence of the intended word. 

Both error types are categorized by the following experimental conditions:
    \begin{itemize}
        \item \textbf{Contextual influence}: Error units may appear in the surrounding linguistic context, as in the bolded words in (1-3), or not (4)
        \item \textbf{Correction status}: Errors may be corrected by the talker, as in (1-3), or not (4)
        \item \textbf{Completion status}: Words may be aborted mid-production (i.e., incomplete), as (2) and (3), or not: (1) and (4)
    \end{itemize}

Contextual errors have the effect of doubling the error term in the local context. This may not have an impact on sound errors, but doubling in word errors may disrupt transcription in infelicitous positions (e.g., \textit{username} in (4)). While corrected errors provide additional evidence for the intended word, they may also confuse ASR models by presenting competing candidates. Incomplete productions, though lacking complete word targets, may benefit ASR performance by reducing evidence for error words and allowing other contextual information to guide transcription. Our experiments explore the impact of these experimental variables on transcription accuracy.

\begin{table}[h]  
    \centering
    \renewcommand{\arraystretch}{1.2}  
    \begin{tabular}{|c|p{0.8\columnwidth}|}  
        \hline
        \textbf{No.} & \textbf{Error Type and Example} \\
        \hline
        1. & \textbf{Contextual Sound Error}: 
        What is it that Republicans don't like /ab\textipa{[I]}t, about \textbf{Mitt} Romney? \\
        \hline
        2. & \textbf{Contextual Word Error}: 
        … but there's a /mov=, book talking about in the future \textbf{movies} \\
        \hline
        3. & \textbf{Incomplete Sound Error}: 
        … the name of this ad is, Facts /Re[k]ar= Regarding their Friable \textbf{Condition} \\
        \hline
        4. & \textbf{Complete Word Error}: 
        You really shouldn't have put WeedLover43 as your uh /username (Intended: password) \\
        \hline
    \end{tabular}
    \caption{Examples of Error Types and Conditions. Notational conventions: /X is an error word, X= is an incomplete word, [X] is a mispronounced sound.}
    \label{tab:errors}
\vspace{-0.5cm}
\end{table}

Speech errors involving sub-lexical sounds are also encoded for the position of the error word and syllable. That is, sound substitutions and deletions are labeled for the following positions: 
    \begin{itemize}
        \item \textbf{Word Position}: Initial, Medial, Final
        \item \textbf{Syllable Position}: Onset, Nucleus, Coda
    \end{itemize}
These categories allow us to examine transcription accuracy relative to sound position.

\subsection{Long-form audio transcription}
Since each podcast episode is 30-60 minutes long, we employed WhisperX \cite{bain2023whisperx} for audio transcription, selected for its non-causal architecture optimized for long-form audio processing. WhisperX combines the use of the Whisper speech models \cite{radford2023robust} with voice activity detection \cite{bredin2023pyannote, silero2024vad} and speaker diarization \cite{bredin2023pyannote}. WhisperX uses wav2vec 2.0 \cite{baevski2020wav2vec} models to force-align Whisper transcriptions for obtaining timestamps and confidence scores. Our transcription setup implemented the \texttt{whisper-large-v2} model with default parameters, including a beam search decoder (beam size=5, patience=1.0). The non-causal architecture enables access to full contextual information, though future research could examine causal models' performance.

\subsection{Data processing and analysis}
The dataset analysis involved several computational steps. WhisperX-generated timestamps and transcriptions were aligned with hand-annotated timestamps and error segments from the SFUSED dataset to extract short-form error representations with precise temporal boundaries. Fuzzy string matching\footnote{https://pypi.org/project/fuzzy-match/} with Levenshtein distance was used for this purpose. The annotated error from the short-form error representation was then checked for an exact string match in the machine generated transcription. We developed an algorithmic classification system for transcription outcomes:

    \begin{itemize}
        \item \textbf{Corrected}: Machine transcription matches the intended (unspoken) word in appropriate context
        \item \textbf{Faithful}: Machine transcription matches the actual spoken error
        \item \textbf{Incorrect}: Cases not fitting either above category, presumed to be machine errors
    \end{itemize}

In the sections below, we isolate contrasts using the specific transcription types above. However, we characterize transcription accuracy in general as the sum of Corrected and Faithful transcriptions, since in both cases the model has accurately transcribed what is spoken (Faithful) or anticipated the intended word (Corrected). 

To improve the classification algorithm, we spot-checked its output with step samples to find misclassifications and then addressed these problems in subsequent versions. After three iterations, the third step sample showed the algorithm had 85\% accuracy. 

While we report the results of our experiments with percentages, in order to examine the relationship between the experimental conditions and transcription type we did chi-square tests of independence on each error type. These tests revealed significant effects across all experimental conditions except contextual sound substitutions and both the word and syllable position effects, though sound deletions and additions were excluded due to insufficient Faithful cases. 

\section{Results}
\vspace{-0.2cm}
WhisperX demonstrated variable transcription accuracy across different error types and conditions. Sound errors achieved 83\% overall accuracy (81\% Corrected, 2\% Faithful, 17\% Incorrect), while word errors showed lower performance at 74\% overall (43\% Corrected, 31\% Faithful, 26\% Incorrect). This disparity aligns with theoretical expectations, given the greater phonetic similarity between intended and error forms in sound errors compared to word errors. All figures in this section use the following color-coding: Blue=Corrected, Yellow=Faithful, Red=Incorrect; bar shading has the following interpretation: Dark=Condition is True, Light=Condition is False, as specified in relevant figures.

\begin{figure}[htbp]
  \centering
  \includegraphics[width=\linewidth, height=0.2\textheight]{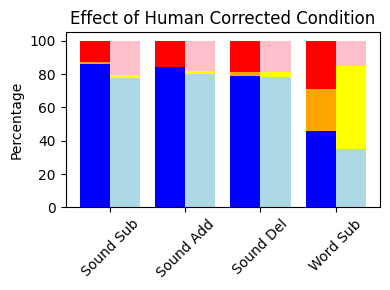}
  \caption{Transcription Types by Corrected Condition: Dark=Corrected, Light=Uncorrected.}
  \label{fig:corrected}
  
  \vspace{0.2em} 
  
  \includegraphics[width=\linewidth, height=0.2\textheight]{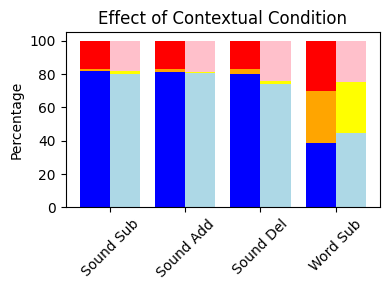}
  \caption{Transcription Types by Contextual Condition: Dark=Contextual, Light=Noncontextual.}
  \label{fig:contextual}
  
  \vspace{0.2em}
  
  \includegraphics[width=\linewidth, height=0.2\textheight]{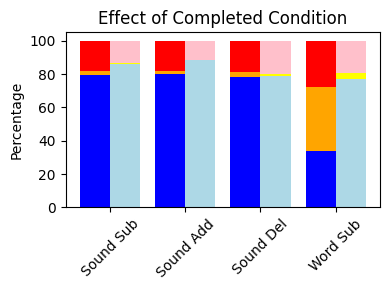}
  \caption{Transcription Types by Completed Condition: Dark=Complete, Light=Incomplete.}
  \label{fig:completed}
  
  \vspace{0.2em}
\end{figure}

Starting first with the Corrected Condition, human correction significantly influenced transcription outcomes across all error categories (Figure~\ref{fig:corrected}). For sound errors, human correction correlated with increased machine-corrected transcriptions and decreased Incorrect/Faithful transcriptions. Word errors exhibited a more complex pattern: while Corrected transcriptions increased (10.91\% improvement), Incorrect transcriptions also rose significantly (13.31\% increase), meaning that human correction decreased overall accuracy in word errors, unlike sound errors. 

The Contextual Condition, which tests for the presence of contextual cues, also produced divergent effects (Figure~\ref{fig:contextual}). Contextual sound errors are transcribed with a marginal improvement in accuracy (1.34\% in Incorrect transcriptions, not significant), particularly pronounced in sound deletions. Word errors, on the other hand, show a difference in the opposite direction: Corrected transcriptions dropped by 6.23\% and Incorrect transcriptions increased by 5.27\%. Thus, reference to the error term in the surrounding context may have a minor positive effect on the transcription of sound errors, but a detrimental effect for word errors. 

The Completed Condition---whether the pronunciation of the speech error is aborted mid-word (as in \textit{Re[k]ar=} for \textit{Regarding})---had a strong negative effect across the board (Figure~\ref{fig:completed}). Incomplete words have higher percentages of Corrected transcriptions in all error types, with a striking 43.13\% difference with word errors. Correspondingly, completed word errors also have much higher percentages of Faithful transcriptions and Incorrect transcriptions, though the difference is smallest with sound deletions. Counterintuitively, incomplete error words yielded higher transcription accuracy despite their degraded acoustic form.

\begin{figure}[htbp]
  \centering
  
  \includegraphics[width=\linewidth, height=0.2\textheight]{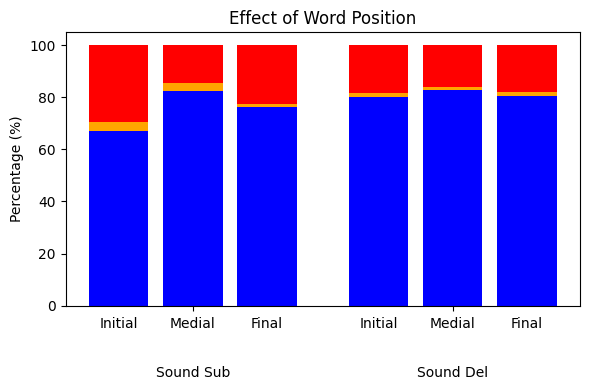}
  \caption{Transcription Types by Word Position.}
  \label{fig:wordposition}
  
  \includegraphics[width=\linewidth]{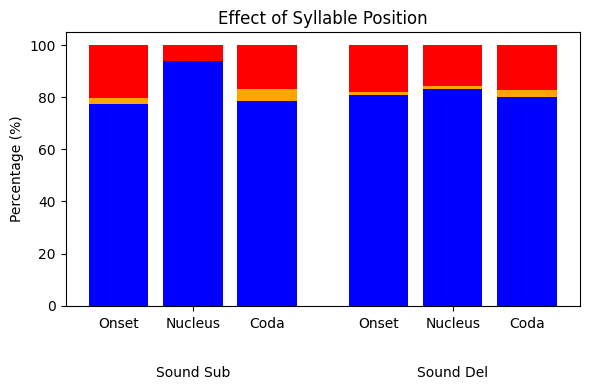}
  \caption{Transcription Types by Syllable Position.}
  \label{fig:syllposition}
\vspace{-0.3cm}
\end{figure}

Finally, we can put the precise location of sound errors under the microscope and examine the impact of location on transcription accuracy (see Figure~\ref{fig:wordposition} and Figure~\ref{fig:syllposition}). Analysis of error position revealed non-significant effects on transcription accuracy, particularly for sound substitutions:
    \begin{itemize}
        \item \textbf{Word position}: Medial $>$ Final $>$ Initial
        \item \textbf{Syllable position}: Nucleus $>$ Coda $>$ Onset
    \end{itemize}
Sound deletions showed less sensitivity to positional effects, suggesting distinct underlying mechanisms for different error types.

\section{Discussion}
A notable finding is the superior transcription accuracy for incomplete error words across all error types (Figure~\ref{fig:completed}). This effect is particularly striking for word errors, where incomplete productions achieve accuracy rates comparable to sound errors (in the 80\% range for Corrected transcriptions). Two mechanisms likely contribute to this phenomenon:
    \begin{itemize}
        \item \textbf{Signal Degradation}: Incomplete production reduces evidence for the error word. This effect is more pronounced in word errors where intended and error words are phonetically distinct. For example, \textit{Re[k]ar=} retains partial information about \textit{Regarding}, while \textit{mov=} (intended: \textit{book}) provides minimal evidence for either word.
        \item \textbf{Contextual Integration}: Word truncation may reduce the model's commitment to specific transcription targets, enabling greater reliance on linguistic context, analogous to cloze tasks in psycholinguistics or masked language modeling in models like BERT \cite{devlin2018bert}. This mechanism is particularly effective for word errors due to their greater phonetic deviation between error and intended forms.
    \end{itemize}

With these explanations in mind, one might wonder why contextual word errors are in general transcribed less accurately (Figure~\ref{fig:contextual}). This decrease in accuracy, despite the presence of error terms in the surrounding context, may be attributed to the nature of repeated elements. While repeated phonemes in sound errors minimally impact transcription, repeated words in contextual word errors likely influence beam search probabilities more substantially. This interference may override richer contextual cues including syntactic, semantic, and argument structure information. This hypothesis is consistent with our conjecture above because the linguistic context for an error is much richer than the existence of a doubled sound or word. Incomplete error words may thus allow for deeper use of this context by avoiding commitment to the wrong word.

Human-corrected errors show increased Corrected transcriptions but decreased overall accuracy for word errors (Figure~\ref{fig:corrected}). This pattern suggests that the presence of both error and correction creates competing transcription targets. The model, lacking awareness of which form is correct, treats both forms as valid candidates. This competition is less problematic for sound errors where error and intended forms are phonetically similar.

Finally, the enhanced accuracy for medial positions in sound errors (word-internal and syllabic nucleus positions) may reflect signal robustness rather than sequential processing advantages. While psycholinguistic models might suggest benefits from sequential prediction \cite{dell1993structure}, WhisperX's non-causal architecture has bidirectional access to context. The superior performance likely stems from the inherent acoustic properties of medial positions, particularly vowels' longer duration and distinct formant structure, which may facilitate error detection and correction.

\section{Conclusion}
Our analysis demonstrates that SFUSED English's cross-classification variables significantly influence ASR transcription accuracy. Sound errors consistently show higher transcription accuracy than word errors, with differential responses to signal degradation conditions (corrected, contextual, completed). While these conditions minimally affect sound errors, they substantially impact word errors, with opposing effects observed in corrected and contextual conditions. Additionally, positional effects within words and syllables specifically influence sound error transcription.

SFUSED English contains several additional unexplored variables that could provide novel ASR evaluation metrics, including phonetic complexity of individual sounds (e.g., fricatives versus simpler sounds), malapropisms (form-related versus unrelated errors), error direction (anticipatory versus perseveratory), and additional error categories (word additions/deletions, morpho-syntactic errors, prosodic errors, etc.). These variables could be particularly valuable in comparing causal versus non-causal models, as speech error structure often depends on access to future linguistic information.

Finally, speech error analysis could enhance named entity recognition, particularly in contextual biasing algorithms in real-world applications like voice search~\cite{wang2023contextual} or transcription of long-form financial earnings calls~\cite{huang-etal-2024-conec}. Contextual biasing refers to the technique of guiding an ASR system to prefer certain words, phrases, or types of vocabulary without retraining. Inference-based approaches often rely on incorporating biasing contexts during the transcription decoding process to enhance decoding scores for specific words or phrases. Understanding common entity-specific error patterns (e.g., \textit{Amazon} → \textit{Amazone}) could enable more sophisticated probability adjustments in keyword recognition systems.   Furthermore, each speaker tends to make speech errors that are typical and characteristic of their speaking style. If a user consistently mispronounces certain words, the system can personalize the biasing. This is especially true with atypical speakers who have characteristic speech error patterns. Our future work looks to investigate contextual biasing algorithms to improve atypical speaker ASR.

\section{Acknowledgements}
The authors would like to acknowledge the Northeastern University Discovery Cluster and Tianyi Zhang for assistance with containerization to help run this work on the cluster. Macarious Hui’s work was supported by a Social Science and Humanities Research Council of Canada grant (435-202-0193), the Khoury West Coast Research Fund (Khoury College of Computer Sciences, Northeastern University), and funding from a 2023 Google Research Scholar Grant entitled ``No speaker left behind: advancing speech technology for disordered speech''.

\bibliographystyle{IEEEtran}
\bibliography{mybib}

\end{document}